%% file: main.tex
\documentclass[10pt,twocolumn,letterpaper]{article}

\usepackage{iccv}              % To produce the CAMERA-READY version

\definecolor{iccvblue}{rgb}{0.21,0.49,0.74}
\usepackage[pagebackref,breaklinks,colorlinks,allcolors=iccvblue]{hyperref}

\usepackage{multirow}
\def\paperID{13} % *** Enter the Paper ID here
\def\confName{ICCV}
\def\confYear{2025}

\title{Optimizing DINOv2 with Registers for Face Anti-Spoofing}

\author{Mika Feng, Pierre Gallin-Martel, Koichi Ito, and Takafumi Aoki\\
Graduate School of Information Sciences, Tohoku University, Japan\\
{\tt\small \{mika, pierre, ito\}@aoki.ecei.tohoku.ac.jp, aoki@ecei.tohoku.ac.jp}
}

\begin{document}
\maketitle
\begin{abstract}
  Face recognition systems are designed to be robust against variations in head pose, illumination, and image blur during capture.
  However, malicious actors can exploit these systems by presenting a face photo of a registered user, potentially bypassing the authentication process.
  Such spoofing attacks must be detected prior to face recognition.
  In this paper, we propose a DINOv2-based spoofing attack detection method to discern minute differences between live and spoofed face images.
  Specifically, we employ DINOv2 with registers to extract generalizable features and to suppress perturbations in the attention mechanism, which enables focused attention on essential and minute features.
  We demonstrate the effectiveness of the proposed method through experiments conducted on the dataset provided by ``The 6th Face Anti-Spoofing Workshop: Unified Physical-Digital Attacks Detection@ICCV2025'' and SiW dataset.  
\end{abstract}

\section{Introduction}
\label{sec:intro}

Face recognition, which identifies individuals based on the shape and position of their facial features, offers significant advantages. 
These include its low cost, requiring only a standard camera for image capture, and its high convenience, enabling contactless and unconstrained identification \cite{Handbook-Face-Recognition}.
The face recognition systems have been designed to be robust against the environmental variations such as head pose, illumination, and image blur during capture, since face images can be easily changed by such variations. 
Consequently, a malicious actor can exploit the face recognition systems by presenting a face photo of a registered user, potentially bypassing the authentication process.
As many face images of individuals are readily available online, such ``spoofing attacks'' have emerged as a realistic threat to face recognition systems \cite{Handbook-Anti-Spoofing}.

Spoofing attacks can be broadly categorized into ``physical attacks'' and ``digital attacks''.
Physical attacks include 2D attacks such as prints, replays, and cutouts, as well as 3D attacks involving masks. 
Digital attacks, on the other hand, encompass digital forgeries such as identity manipulation, adversarial examples, and generated images by Generative Adversarial Networks (GAN) \cite{Goodfellow-CACM-2020}.
As discussed, despite the inherent robustness of face recognition systems to environmental variations, it is crucial to detect these spoofing attacks prior to the authentication process to ensure system security.

To effectively counter spoofing attacks, it is essential to detect minute differences between live and spoofed face images.
It is critical to extract features from the input image that reveal global or local discrepancies, such as the texture of paper, a display device,
% reflection, interference fringes of display device,
depth, etc.
Many methods \cite{Yu-CVPR-2020,Yu-PAMI-2020,Wang-CVPR-2022,Watanabe-APSIPA-2022,Chen-CoRR-2023,Wang-TBIOM-2022,Li-NN-2024,Zheng-IFS-2024,Feng-CVPRW-2025,He-CVPRW-2024} utilizing Convolutional Neural Networks (CNNs) and Vision Transformers (ViT) \cite{Dosovitskiy-ICLR-2021} have been proposed to extract features inherent to spoofing attacks.
Although these methods achieve high detection accuracy for simple spoofing attacks, their performance tends to degrade significantly when faced with unknown spoofing attacks not included in the training data.

In this paper, we propose a novel spoofing attack detection method utilizing DINOv2 \cite{Oquab-TMLR-2024}, enabling the capture of fine-grained details within a face image.
We employ DINOv2 with registers \cite{Darcet-ICLR-2024} to enhance the generalizability of extracted features and to suppress perturbations in the attention mechanism, thereby enabling focused attention on essential and minute features.
Furthermore, to maintain the generalization capability of the extracted features, only the last encoder block of DINOv2 is made trainable.
We demonstrate the effectiveness of the proposed method through extensive experiments conducted on the dataset provided by ``The 6th Face Anti-Spoofing Workshop: Unified Physical-Digital Attacks Detection@ICCV2025'' \cite{Liu-CoRR-2025}, hereafter referred to as the FAS Workshop dataset, and the SiW dataset \cite{Liu-CVPR-2018}.

\begin{figure*}[t]
  \centering
  \includegraphics[width=\linewidth]{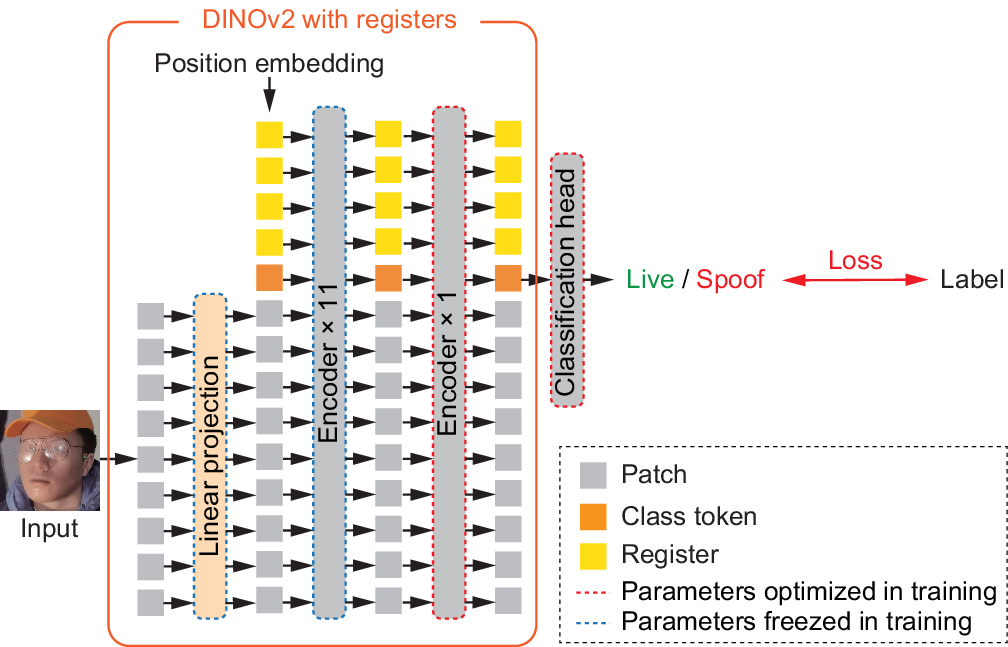}
  \caption{Overview of the proposed method.}
  \label{fig:proposed}
\end{figure*}

\section{Related Work}

This section presents a brief overview of face spoofing attack detection methods proposed with intra-dataset testing.
% \subsection{Face Spoofing Attack Detection}
Major methods for detecting face spoofing attacks utilize CNNs and ViT.
There are several CNN-based methods such as CDCN++ \cite{Yu-CVPR-2020}, NAS-FAS \cite{Yu-PAMI-2020}, and PatchNet \cite{Wang-CVPR-2022}.
CDCN++ \cite{Yu-CVPR-2020} extracts depth features of face images using CNN and detects spoofing attacks using these features.
NAS-FAS \cite{Yu-PAMI-2020} detects spoofing attacks using an optimal network derived from Neural Architecture Search (NAS).
PatchNet \cite{Wang-CVPR-2022} employs a patch-based approach, dividing face images into patches and extracting local features from each patch using CNN to detect spoofing attacks.
CNNs cannot always recognize the various changes that occur in a face image due to spoofing attacks.
In particular, the detection accuracy is degraded for unknown spoofing attacks that are not included in the training data.
ViT \cite{Dosovitskiy-ICLR-2021} can extract both local and global features from an image, and thus can detect local and global changes in a face image caused by spoofing attacks.
TransFAS \cite{Wang-TBIOM-2022} improved the accuracy of spoofing attack detection by adding a depth feature extraction module to ViT.
Watanabe et al. \cite{Watanabe-APSIPA-2022} introduced the data augmentation method, Patch-wise Data Augmentation (PDA), which is dedicated to detecting spoofing attacks, into ViT to improve the accuracy of spoofing attack detection using ViT.
Chen et al. \cite{Chen-CoRR-2023} proposed a spoofing attack detection method using Segment Anything Model (SAM) \cite{Kirillov-ICCV-2023}, which is a foundation model of image segmentation based on ViT.
Feng et al. \cite{Feng-CVPRW-2025} proposed a spoofing attack detection method that uses the features extracted from the intermediate layers of ViT and employs data augmentation methods enhancing the performance of face spoofing attack detection: Patch-wise Data Augmentation (PDA) \cite{Watanabe-APSIPA-2022} and Face Anti-Spoofing data augmentation (FAS-Aug) \cite{Cai-IJCV-2024}.
% Watanabe et al. \cite{Watanabe-APSIPA-2022} and Chen et al. \cite{Chen-CoRR-2023} utilize only features extracted from the deep layer of ViT to detect spoofing attacks, and thus do not consider local and global features in an optimal balance.
% TransFAS \cite{Wang-TBIOM-2022} utilizes intermediate features of ViT, however, it has not been sufficiently investigated whether it necessarily uses features extracted from the suitable layer for detecting spoofing attacks.

% \begin{figure*}[t]
%   \centering
%   \includegraphics[width=\linewidth]{proposed.eps}
%   \caption{Overview of the proposed method.}
%   \label{fig:proposed}
% \end{figure*}

\section{Proposed Method}

We describe a face spoofing attack detection method that utilizes DINOv2 with registers proposed in this paper.

% \begin{figure*}[t]
%   \centering
%   \includegraphics[width=\linewidth]{proposed.eps}
%   \caption{Overview of the proposed method.}
%   \label{fig:proposed}
% \end{figure*}

\subsection{Overview}
\label{sec:overview}

Fig. \ref{fig:proposed} illustrates the network architecture based on DINOv2 ViT-B/14 with registers \cite{Darcet-ICLR-2024}.
This architecture takes an input image of $224 \times 224$ pixels, divides it into $14 \times 14$ patches, and feeds them into an encoder consisting of 12 layers.
We use pre-trained parameters as initial values and then fine-tune the model on the training data of a face spoofing attack detection dataset.
To prevent overfitting on the training data and maintain generalization performance, we unfreeze and fine-tune only the parameters of the last encoder block of DINOv2.
The effectiveness of this partial unfreezing strategy for the last encoder block has been empirically confirmed, with further details provided in Sect. \ref{sec:ablation_unfreeze}.
A classification head is appended after the last layer of DINOv2 with registers.
For training, we utilize a 2-class classification loss applied to the class token output from the classification head.
For inference, the live probability is calculated from the output of the classification head.

\subsection{DINOv2 with Registers}
\label{sec:DINOv2}

The proposed method employs DINOv2 with registers to effectively capture the subtle and intrinsic differences between live and spoofed images.
In detecting spoofing attacks, the differences in fine textures and patterns between live and spoofed images directly affect the detection performance.
Consequently, if noise appears in the attention map of ViT, it may prevent ViT from learning of essential and fine-grained features indispensable for effective spoofing attack detection.
Prior research visualizing the internal operations of ViT has observed a ``spike phenomenon'' where abnormally high attention is focused on non-essential patches, such as background regions \cite{Darcet-ICLR-2024}.
Such perturbations in the attention maps is attributed to outlier tokens within a portion of the embeddings output from the model, which exhibit extremely high vector norms.
It has been reported that they contain global information about the entire image, but lack local features.
This phenomenon is considered to be due to the reuse of unnecessary patches as temporary memory storage in ViT.
To address this issue, we introduce register tokens \cite{Darcet-ICLR-2024} in ViT.
A register is a learnable token that is treated as part of the input sequence, similar to patch and class tokens in Transformer \cite{Dosovitskiy-ICLR-2021}.
In DINOv2 ViT-B/14 with registers, four register tokens are introduced to provide a structure dedicated to the retention and aggregation of temporary information inside the model, thereby suppressing attention disturbances caused by the reuse of unnecessary patches.
Thus, by using DINOv2 with registers, the method suppresses perturbations in the attention mechanism and allows for focused attention on essential and fine-grained features.

\subsection{Training Strategy}

We describe the training strategy employed for our proposed face anti-spoofing model.
To effectively train our model for robust face anti-spoofing, we employ the AdamW optimizer with differentiated learning rate: $5 \times 10^{-5}$ for the binary classification head and a smaller rate of $5 \times 10^{-6}$ for the DINOv2 backbone.
This separation was motivated by the need to fine-tune the high-level classifier more aggressively, while preserving the pre-trained representational power of the backbone by applying smaller updates.
We use a cosine annealing scheduler to smoothly reduce the learning rate over time, helping stabilize training and improve convergence.
To address the disparity in difficulty between attack types, we employ the focal loss function with a gamma value of 2 and equal class weights.
This places greater emphasis on hard-to-classify examples by down-weighting the loss contribution of well-classified samples, thereby encouraging the model to focus on more informative and challenging inputs.
The use of equal class weights ensures that both the live and spoof categories are treated symmetrically during optimization.

\begin{figure*}[t]
  \centering
  \includegraphics[width=\linewidth]{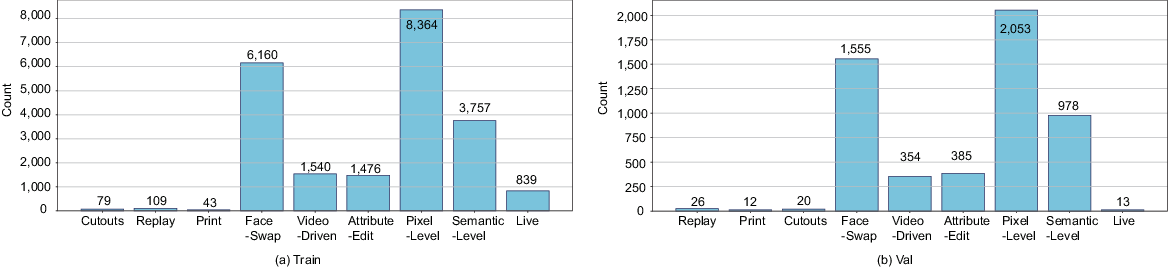}
  \caption{Distribution of the classes in the datasets provided from ``The 6th Face Anti-Spoofing Workshop: Unified Physical-Digital Attacks Detection@ICCV2025'': (a) training dataset and (b) validation dataset.}
  \label{fig:dist}
\end{figure*}

\section{Experiments and Discussion}

This section describes the experiments conducted to demonstrate the effectiveness of the proposed method in detecting face spoofing attacks.
The experiments use two distinct datasets to comprehensively evaluate the performance.
First, we use the FAS Workshop dataset \cite{Liu-CoRR-2025} provided by ``The 6th Face Anti-Spoofing Workshop: Unified Physical-Digital Attacks Detection@ICCV2025.''
Second, we use the publicly available SiW dataset \cite{Liu-CVPR-2018} to allow for a comparison with existing methods.
The following discussion is divided into two parts, with each section dedicated to one of these datasets.

\subsection{Experiments on FAS Workshop Dataset}

This part focuses on the experiments conducted on the FAS Workshop dataset. 

\subsubsection{Dataset}

The FAS Workshop dataset consists of a variety of spoofing attacks: ``Live'', ``Print'', ``Replay'', ``Cutouts'', ``Transparent'', ``Plaster'', ``Resin'', ``Attribute-Edit'', ``Face-Swap'', ``Video-Driven'', ``Pixel-Level'', ``Semantic-Level'', ``ID Consistent'', ``Style Transfer'', and ``Prompt Driven''.
The detailed description of this dataset is available in \cite{Liu-CoRR-2025}.
It is an enhanced version of UniAttackData \cite{Fang-IJCAI-2024}, which includes datasets such as \cite{Zhang-CVPR-2019, Zhang-TBIOM-2020, Liu-WACV-2021, Liu-IFS-2022}.
Fig. \ref{fig:dist} shows the distribution of the classes in training dataset and validation dataset, revealing a strong class imbalance.
% OULU-NPU consists of 4,950 videos taken from 55 subjects.
% Each frame is $1,080 \times 1,920$ pixels and each video is captured at 30 fps for about 15 seconds.
% The ``Live'' videos are captured using the front cameras of 6 mobile devices, i.e., Samsung Galaxy S6 edge, HTC Desire EYE, MEIZU X5, ASUS Zenfone Selfie, Sony XPERIA C5 Ultra Dual, and OPPO N3, in 3 sessions with different illumination conditions and background scenes
% The print attack uses a paper printed by two types of printers, and the display attack uses two types of display devices.

\subsubsection{Experimental Condition}
\label{sec:FAS-ExpCon}

We describe the experimental conditions in the experiments.
The input images are normalized with training dataset statistical characteristics, namely average and standard deviation.
The image is divided into patches and input to the proposed method.
As described in Sect. \ref{sec:overview}, the proposed method uses the pre-trained model of DINOv2 ViT-B/14 with registers \cite{Darcet-ICLR-2024} and performs fine tuning on the FAS Workshop dataset.
The size of the mini-batch is 32, the learning rate for the binary classification head is $5 \times 10^{-5}$, the learning rate for the DINOv2 backbone is $5 \times 10^{-6}$, the number of epochs is 200 with early stopping with a patience of 20 epochs, and AdamW is used as optimizer.
We conduct two experiments to demonstrate the effectiveness of the proposed method.
The first experiment is the ablation study on unfreezing layers for DINOv2 with registers (Sect. \ref{sec:ablation_unfreeze}).
The second experiment is a comparison between the baseline method (AjianLiu) in ``The 6th Face Anti-Spoofing Workshop: Unified Physical-Digital Attacks Detection@ICCV2025'' and the proposed method (Sect. \ref{sec:comparison_challenge}).

\subsubsection{Evaluation Metrics}

In this experiment, we use Attack Presentation Classification Error Rate (APCER), Bonafide Presentation Classification Error Rate (BPCER), Average Classification Error Rate (ACER), Area Under the receiver operating characteristic Curve (AUC), Accuracy (ACC), and Equal Error Rate (EER).
APCER measures the false acceptance rate for spoofing attacks and is calculated by
\begin{equation}
  \mathrm{APCER} = 1 - \frac{1}{N_{PA}} \sum_{i=1}^{N_{PA}} R_i,
\end{equation}
where $PA$ denotes the set of spoofing attack samples, $N_{PA}$ is the number of spoofed images, and $R_i$ is a binary indicator function that takes 1 if the image is classified as ``Spoof'' and 0 if classified as ``Live.''
BPCER is the false rejection rate for bonafide (live) images, defined as
\begin{equation}
  \mathrm{BPCER} = \frac{1}{N_{BF}} \sum_{i=1}^{N_{BF}} R_i,
\end{equation}
where $N_{BF}$ is the number of live images.
ACER is defined by the average error as
\begin{equation}
  \mathrm{ACER} = \frac{\mathrm{APCER} + \mathrm{BPCER}}{2}.
\end{equation}
AUC reflects the ability to distinguish between live and spoof samples across various thresholds; higher AUC values indicate better overall separability.
ACC measures the overall proportion of correctly classified images both live and spoof and is defined as the ratio of correct predictions to the total number of samples.
EER is the error rate for a given threshold, where the false acceptance rate (FAR) equals the false rejection rate (FRR); lower EER indicates better balance between detecting spoof attempts and avoiding misclassification of live images.
The smaller the values of APCER, BPCER, ACER, and EER, and the higher the values of AUC and ACC, the better the model's spoof detection performance.

\subsubsection{Exp. (i): Unfreezing layers for DINOv2 with registers}
\label{sec:ablation_unfreeze}

In this experiment, we evaluate the effectiveness of unfreezing only the last encoder block of DINOv2 with registers for training.
Table \ref{tbl:Ablation_1} shows the experimental results, where the highest accuracy is shown in bold, and the second highest accuracy is underlined.
We observe that reducing the number of trainable layers leads to lower accuracy on the validation dataset but higher accuracy on the test dataset.
This suggests a significant distribution shift between the validation dataset and the test dataset.
Since unfreezing only the last encoder block exhibits the best on the test dataset, all subsequent experiments with the proposed method will adopt this strategy.

\begin{table}[t]
  \centering
  \caption{Experimental results of unfreezing layers for DINOv2 with registers, where the highest accuracy is shown in bold, and the second highest accuracy is underlined.}
  \label{tbl:Ablation_1}
  \begin{tabular}{@{}lcccc@{}}
    \toprule
    Unfreezed & \multicolumn{3}{c}{Validation} & Test \\
    \cmidrule(lr){2-4} \cmidrule(l){5-5}
    layers      & APCER $\downarrow$ & BPCER $\downarrow$ & ACER $\downarrow$ & ACER $\downarrow$ \\
    \midrule
    10$\sim$12 & \textbf{0.4334} & \underline{0.0769} & \textbf{0.2552} & 0.2147 \\
    11$\sim$12 & \underline{0.575}  & \textbf{0.0}       & \underline{0.2875}    & \underline{0.1443} \\
    12             & 0.6318          & \textbf{0.0}       & 0.3159          & \textbf{0.1107} \\
    \bottomrule
  \end{tabular}
\end{table}

\subsubsection{Exp. (ii): Comparison between the Baseline Method and the Proposed Method}
\label{sec:comparison_challenge}

In this experiment, we compare the proposed method with the baseline method (AjianLiu) in ``The 6th Face Anti-Spoofing Workshop: Unified Physical-Digital Attacks Detection@ICCV2025''.
Table \ref{tbl:Proposed_vs_Baseline} shows the experimental results, where the highest accuracy is shown in bold.
Our proposed method performs better detection accuracy than the baseline method on ACER, AUC, and ACC, while the baseline method exhibits better performance on EER.
This result indicates that the generalization performance of the proposed method is better than the baseline method against unknown spoofing attacks, while there is room for performance improvement.
The reason for the poor performance of the proposed method in terms of EER (0.6132 vs. baseline 0.2307) is an imbalance between APCER and BPCER.
Since EER is defined as the point where the false acceptance rate (related to APCER) equals the false rejection rate (related to BPCER), a large gap between these two metrics makes it difficult to set a threshold that performs well for both.
As a result, the threshold must be placed at a position that is suboptimal for each metric, leading to a higher EER.

\begin{table}[t]
  \centering
  \caption{Comparison of the baseline method in the FAS workshop and the proposed method using the provided dataset, where the highest accuracy is shown in bold.}
  \label{tbl:Proposed_vs_Baseline}
  \begin{tabular}{@{}lcccc@{}}
    \toprule
    Method & ACER $\downarrow$ & AUC $\uparrow$ & ACC $\uparrow$ & EER $\downarrow$ \\
    \midrule
    Baseline (AjianLiu) & 0.2259 & 0.8989 & 0.6355 & \bf 0.2307 \\
    Proposed & \bf 0.1107 & \bf 0.9480 & \bf 0.9047 & 0.6132 \\
    \bottomrule
  \end{tabular}
\end{table}

\subsection{Experiments on SiW Dataset}

This part focuses on the experiments conducted on the SiW dataset. 

\subsubsection{Dataset}

SiW consists of 4,478 videos taken from 165 subjects.
For each subject, there are 8 real videos and up to 20 spoofed videos.
Each frame is $1,920 \times 1,080$ pixels and each video is captured at 30 fps for about 15 seconds.
The ``Live'' videos are captured under various lighting conditions while the subjects move their heads back and forward, rotate their heads, and change their facial expressions.
The print attack uses two types of paper, and the display attack uses four types of display devices.
For a detailed explanation of Protocols of SiW dataset, please refer to \cite{Liu-CVPR-2018}.

\subsubsection{Data Augmentation}

\begin{figure}[t]
  \centering
  \includegraphics[width=\linewidth]{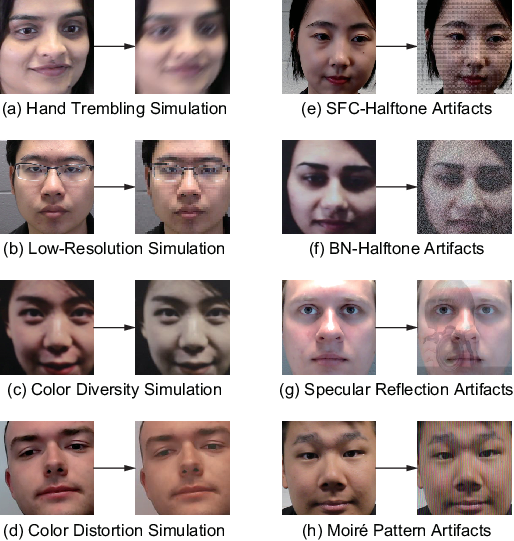}
  \caption{Example of face images after applying FAS-Aug.}
  \label{fig:FAS-Aug}
\end{figure}

Since SiW only includes physical attack types, augmentation techniques can be used to great effect during training.
In this experiment, we employ two data augmentation methods: a traditional method and FAS-Aug \cite{Cai-IJCV-2024}.
The traditional data augmentation method refers to simple, non FAS-specific alterations such as random rotation and color jitter.
FAS-Aug \cite{Cai-IJCV-2024} consists of 8 types of data augmentation that emulate photography noise, print attack, and display attack.
As photography noise, (a) hand trembling simulation, (b) low-resolution simulation, and (c) color diversity simulation are added to the image.
Since these noises occur in both live and spoofed images, the labels are not replaced when these noises are added.
As print-attack noise, (d) color distortion simulation, (e) SFC-halftone artifacts, and (f) BN-halftone artifacts are added to the image.
When these are added, the label is replaced by ``Spoof.''
As display-attack noise, (g) specular reflection artifacts and (h) moir\'{e} pattern artifacts are added to the image.
When these are added, the label is replaced by ``Spoof.''
The parameters used in FAS-Aug are in accordance with \cite{Cai-IJCV-2024}.
The proposed method randomly selects one of the 8 data augmentations and applies it to the input image.
Fig. \ref{fig:FAS-Aug} shows the example of images to which the 8 data augmentations included in FAS-Aug are applied.

\subsubsection{Experimental Condition}

The images in SiW need to be processed before conducting the experiments.
Since videos in SiW \cite{Liu-CVPR-2018} contain objects that can easily be recognized as spoofing attacks, face regions are extracted from each frame of the videos and used as input images.
% For OULU-NPU \cite{Boulkenafet-FG-2017}, face regions are extracted from each frame using MTCNN \cite{Xiang-ICISCE-2017} and resized to $224 \times 224$ pixels.
Since the bounding box for face regions is known, the face regions are extracted based on these coordinates and resized to $224 \times 224$ pixels.
The proposed method assumes that the input is a single image, not a video.
Therefore, 5 frames are randomly extracted from the training or evaluation videos for each evaluation protocol of SiW, and used as input images for training or evaluation in the experiments as in \cite{Boulkenafet-FG-2017}.
Similarly to Sect. \ref{sec:FAS-ExpCon}, the input images are normalized with training dataset characteristics, and the same model and parameters are used.
However, we made a few adjustments due to the characteristics of SiW, which has less diverse and challenging attack types than the FAS Workshop dataset.
We unfreeze the entire model for training, and switch the loss function to cross entropy and the optimizer to Nesterov Stochastic Gradient Descent \cite{Nesterov-USSR-1983}.
For data augmentation, we systematically apply a traditional method, and use FAS-Aug with a probability of 0.2.
We conduct a comparison between conventional methods and the proposed method to evaluate the effectiveness of the proposed method in detecting face spoofing attacks (Sect. \ref{sec:comparison}).

\subsubsection{Evaluation Metrics}

For this experiment, we use three evaluation metrics: APCER, BPCER, and ACER.
We adopt a more rigorous definition of APCER that considers only the worst-case error rate across all attack types, rather than their average.
This approach enables a fairer comparison with conventional methods from the SiW experiment, which also adopt this formula.

\subsubsection{Exp. (iii): Comparison between Conventional Methods and the Proposed Method}
\label{sec:comparison}

In this experiment, we compare the proposed method with conventional methods for detecting face spoofing attacks.
We used the following conventional methods: CDCN++ \cite{Yu-CVPR-2020}, NAS-FAS \cite{Yu-PAMI-2020}, PatchNet \cite{Wang-CVPR-2022}, Watanabe \cite{Watanabe-APSIPA-2022}, TransFAS \cite{Wang-TBIOM-2022}, MFAE \cite{Zheng-IFS-2024}, Li \cite{Li-NN-2024}, and Feng \cite{Feng-CVPRW-2025}.
Table \ref{tbl:results_SiW} shows the experimental results for SiW, where the highest accuracy is shown in bold, and the second highest accuracy is underlined.
In Protocol 1 and 2, all the methods exhibit low values of the evaluation metrics, and therefore, can detect spoofing attacks with high accuracy.
In other words, Protocol 1 and 2 are easy evaluation protocols.
Protocol 3, on the other hand, evaluates the detection accuracy of unknown spoofing attacks and thus the value of the evaluation metrics becomes higher.
The lowest ACER of the conventional methods is 0.83$\pm$0.13\%.
ACER of Proposed is 0.625$\pm$0.495\%, which is significantly lower than that of the conventional methods.
These results demonstrate the effectiveness of leveraging DINOv2 with registers and introducing data augmentation specialized for spoofing attack detection.

\begin{table}[t]
  \centering
  \caption{Experimental results of each method for SiW (Unit: \%).}
  \label{tbl:results_SiW}
  \scalebox{0.8}[0.8]{
    \begin{tabular}{ccccc}
    \hline
    Prot. & Method & APCER$\downarrow$ & BPCER$\downarrow$ & ACER$\downarrow$ \\
    \hline
    \multirow{9}{*}{1} 
     & CDCN++ \cite{Yu-CVPR-2020} & \underline{0.07} & 0.17 & 0.12 \\ 
     & NAS-FAS \cite{Yu-PAMI-2020} & \underline{0.07} & 0.17 & 0.12 \\ 
     & PatchNet \cite{Wang-CVPR-2022} & {\bf 0.00} & {\bf 0.00} & {\bf 0.00} \\ 
     & Watanabe \cite{Watanabe-APSIPA-2022} & 0.11 & \underline{0.08} & 0.10 \\ 
     & TransFAS \cite{Wang-TBIOM-2022} & {\bf 0.00} & {\bf 0.00} & {\bf 0.00} \\
     & MFAE \cite{Zheng-IFS-2024} & {\bf 0.00} & {\bf 0.00} & {\bf 0.00} \\
     & Li \cite{Li-NN-2024} & {\bf 0.00} & 0.16 & \underline{0.08} \\
     & Feng \cite{Feng-CVPRW-2025} & 0.1 & \underline{0.08} & 0.09 \\
     & Proposed & {\bf 0.00} & {\bf 0.00} & {\bf 0.00} \\
     \hline
    \multirow{9}{*}{2} 
     & CDCN++ \cite{Yu-CVPR-2020} & {\bf 0.00$\pm$0.00} & 0.09$\pm$0.10 & 0.04$\pm$0.05 \\ 
     & NAS-FAS \cite{Yu-PAMI-2020} & {\bf 0.00$\pm$0.00} & 0.09$\pm$0.10 & 0.04$\pm$0.05 \\ 
     & PatchNet \cite{Wang-CVPR-2022} & {\bf 0.00$\pm$0.00} & {\bf 0.00$\pm$0.00} & {\bf 0.00$\pm$0.00} \\ 
     & Watanabe \cite{Watanabe-APSIPA-2022} & \underline{0.01$\pm$0.01} & \underline{0.01$\pm$0.01} & \underline{0.01$\pm$0.01} \\
     & TransFAS \cite{Wang-TBIOM-2022} & {\bf 0.00$\pm$0.00} & {\bf 0.00$\pm$0.00} & {\bf 0.00$\pm$0.00} \\
     & MFAE \cite{Zheng-IFS-2024} & {\bf 0.00$\pm$0.00} & {\bf 0.00$\pm$0.00} & {\bf 0.00$\pm$0.00} \\
     & Li \cite{Li-NN-2024} & {\bf 0.00$\pm$0.00} & 0.16$\pm$0.00 & 0.08$\pm$0.00 \\
     & Feng \cite{Feng-CVPRW-2025} & 0.02$\pm$0.03 & 0.02$\pm$0.03 & 0.02$\pm$0.03 \\
     & Proposed & {\bf 0.00$\pm$0.00} & {\bf 0.00$\pm$0.00} & {\bf 0.00$\pm$0.00} \\
     \hline
    \multirow{9}{*}{3} 
     & CDCN++ \cite{Yu-CVPR-2020} & 1.97$\pm$0.33 & 1.77$\pm$0.10 & 1.90$\pm$0.15 \\ 
     & NAS-FAS \cite{Yu-PAMI-2020} & 1.58$\pm$0.23& 1.46$\pm$0.08 & 1.52$\pm$0.13 \\ 
     & PatchNet \cite{Wang-CVPR-2022} & 3.06$\pm$1.1 & 1.83$\pm$0.83 & 2.45$\pm$0.45 \\ 
     & Watanabe \cite{Watanabe-APSIPA-2022} & 3.07$\pm$2.75 & 3.07$\pm$2.75 & 3.07$\pm$2.75 \\
     & TransFAS \cite{Wang-TBIOM-2022} & 1.95$\pm$0.40 & 1.92$\pm$0.11 & 1.94$\pm$0.26 \\
     & MFAE \cite{Zheng-IFS-2024} & 2.57$\pm$1.83 & 1.92$\pm$1.06 & 2.42$\pm$1.45 \\
     & Li \cite{Li-NN-2024} & 2.13$\pm$1.22 & 2.25$\pm$1.06 & 2.19$\pm$1.14 \\
     & Feng \cite{Feng-CVPRW-2025} & \underline{0.83$\pm$0.13} & \underline{0.84$\pm$0.14} & \underline{0.83$\pm$0.13} \\
     & Proposed & {\bf 0.625$\pm$0.565} & {\bf 0.585$\pm$0.455} & {\bf 0.625$\pm$0.495} \\
     \hline
    \end{tabular}
  }
\end{table}

\section{Conclusion}

In this paper, we proposed a novel spoofing attack detection method utilizing DINOv2 with registers to effectively capture minute and intrinsic differences between live and spoofed face images.
Our approach leverages the ability of DINOv2 to extract fine-grained details, crucial for robust anti-spoofing.
The incorporation of register tokens within DINOv2 significantly enhances the generalizability of extracted features and suppresses attention mechanism perturbations, enabling focused attention on essential features.
To further maintain this generalization, only the last encoder block of DINOv2 was made trainable.
We demonstrated the effectiveness of our method through the experiments on the FAS Workshop dataset and the SiW dataset, confirming its potential to advance face anti-spoofing capabilities.

\section{Acknowledgment}

This work was supported in part by JSPS KAKENHI JP 23H00463 and 25K03131.

{
    \small
    \bibliographystyle{ieeenat_fullname}
    \bibliography{main}
}

\end{document}

%% file: main.bib
@String(PAMI = {IEEE Trans. Pattern Anal. Mach. Intell.})

@String(PAMI  = {IEEE TPAMI})

@book{Handbook-Anti-Spoofing,
   author       = "Marcel, S. and Fierrez, J. and Evans, N.",
   title        = "Handbook of Biometric Anti-Spoofing",
   publisher    = "Springer",
   year         = "2023"
}

@book{Handbook-Face-Recognition,
   author       = "Li, S. and Jain, A.",
   title        = "Handbook of Face Recognition",
   publisher    = "Springer",
   year         = "2011"
}

@article{Yu-CVPR-2020,
  author  = "Yu, Z. and Zhao, C. and Wang, Z. and Qin, Y. and Su, Z. and Li, X. and Zhou, F. and Zhao, G.",
  title   = "Searching central difference convolutional networks for face anti-spoofing",
  journal = "Proc. IEEE/CVF Conf. Comput. Vis. Pattern Recog.",
  pages   = "5295--5305",
  year    = "2020",
  month   = mar
}

@article{Watanabe-APSIPA-2022,
  author  = "Watanabe, K. and Ito, K. and Aoki, T.",
  title   = "Spoofing attack detection in face recognition system using vision transformer with patch-wise data augmentation",
  journal = "Proc. Asia-Pacific Signal and Information Processing Association Annual Summit and Conf.",
  pages   = "1561--1565",
  year    = "2022",
  month   = nov
}

@article{Wang-TBIOM-2022,
  author  = "Wang, Z. and Wang, Q. and Deng, W. and Guo, G.",
  title   = "Face anti-spoofing using transformers with relation-aware mechanism",
  journal = "IEEE Trans. Biometrics, Behavior, and Identity Science",
  volume  = "4",
  number  = "3",
  pages   = "439--450",
  year    = "2022",
  month   = jul
}

@article{Dosovitskiy-ICLR-2021,
  author  = "Dosovitskiy, A. and Beyer, L. and Kolesnikov, A. and Weissenborn, D. and Zhai, X. and Unterthiner, T. and Dehghani, M. and Minderer, M. and Heigold, G. and Gelly, S. and Uszkoreit, J. and Houlsby, N.",
  title   = "An image is worth 16x16 words: {T}ransformers for image recognition at scale",
  journal = "Proc. Int'l Conf. Learn. Represent.",
  year    = "2021",
  month   = jan
}

@article{Liu-CVPR-2018,
  author  = "Liu, Y. and Jourabloo, A. and Liu, X.",
  title   = "Learning deep models for face anti-spoofing: {B}inary or auxiliary supervision",
  journal = "Proc. IEEE/CVF Conf. Comput. Vis. Pattern Recog.",
  pages   = "389--398",
  year    = "2018",
  month   = jun
}

@article{Boulkenafet-FG-2017,
  author  = "Boulkenafet, Z. and Komulainen, J. and Li, L. and Feng, X. and Hadid, A.",
  title   = "{OULU-NPU}: {A} mobile face presentation attack database with real-world variations",
  journal = "Proc. IEEE Int'l Conf. Automatic Face Gesture Recog.",
  pages   = "612--618",
  year    = "2017",
  month   = jun
}

@article{Nesterov-USSR-1983,
  author  = "Nesterov, Y.",
  title   = "A method of solving a convex programming problem with convergence rate ${O}(1/k^2)$",
  journal = "Proc. USSR Academy of Sciences",
  volume  = "269",
  pages   = "543--547",
  year    = "1983",
  month   = may
}

@article{Yu-PAMI-2020,
  author  = "Yu, Z. and Wan, J. and Qin, Y. and Li, X. and Li, S. Z. and Zhao, G.",
  title   = "{NAS-FAS}: {S}tatic-dynamic central difference network search for face anti-spoofing",
  journal = PAMI,
  volume  = "43",
  number  = "9",
  pages   = "3005--3023",
  year    = "2020",
  month   = nov
}

@article{Wang-CVPR-2022,
  author  = "Wang, C. and Lu, Y. and Yang, S. and Lai, S.",
  title   = "Patch{N}et: {A} simple face anti-spoofing framework via fine-grained patch recognition",
  journal = "Proc. IEEE/CVF Conf. Comput. Vis. Pattern Recog.",
  pages   = "20281--20290",
  year    = "2022",
  month   = jun
}

@article{Chen-CoRR-2023,
  author  = "Chen, X. and Jia, Y. and Wu, Y.",
  title   = "Fine-grained annotation for face anti-spoofing",
  journal = "CoRR",
  volume  = "abs/2310.08142",
  year    = "2023",
  month   = oct
}

@article{Kirillov-ICCV-2023,
  author  = "Kirillov, A. and Mintun, E. and Ravi, N. and Mao, H. and Rolland, C. and Gustafson, L. and Xiao, T. and Whitehead, S. and Berg, A. C. and Lo, W.-Y. and Girshick, R.",
  title   = "Segment anything",
  journal = "Proc. IEEE/CVF Int'l Conf. Comput. Vis.",
  pages   = "4015--4026",
  year    = "2023",
  month   = oct
}

@article{Cai-IJCV-2024,
  author  = "Cai, R. and Soh, C. and Yu, Z. and Li, H. and Yang, W. and Kot, A.",
  title   = "Towards data-centric face anti-spoofing: {I}mproving cross-domain generalization via physics-based data synthesis",
  journal = "Int'l J. Comput. Vis.",
  pages   = "1--22",
  year    = "2024",
  month   = oct
}

@article{Li-NN-2024,
  author  = "Li, D. and Chen, G. and Wu, X. and Yu, Z. and Tan, M.",
  title   = "Face anti-spoofing with cross-stage relation enhancement and spoof material perception",
  journal = "Neural Networks",
  volume  = "175",
  pages   = "106275",
  year    = "2024",
  month   = jul
}

@article{Zheng-IFS-2024,
  author  = "Zheng, T. and Li, B. and Wu, S. and Wan, B. and Mu, G. and Liu, S. and Ding, S. and Wang, J.",
  title   = "{MFAE}: {M}asked frequency autoencoders for domain generalization face anti-spoofing",
  journal = "IEEE Trans. Inf. Forensics Secur.",
  pages   = "4058--4069",
  year    = "2024",
  month   = feb
}

@article{Darcet-ICLR-2024,
  author  = "Darcet, T. and Oquab, M. and Mairal, J. and Bojanowski, P.",
  title   = "Vision transformer needs register",
  journal = "Proc. Int'l Conf. Learn. Represent.",
  year    = "2024",
  month   = apr
}

@article{Liu-CoRR-2025,
  author  = "Liu, A. and Yuan, H. and Guo, X. and Ma, H. and Zhuang, W. and Miao, C. and Hong, Y. and Song, C. and Lan, J. and Chu, Q. and Gong, T. and Liang, Y. and Wang, W. and Wan, J. and Liu, X. and Lei, Z.",
  title   = "Benchmarking unified face attack detection via hierarchical prompt tuning",
  journal = "CoRR",
  volume  = "abs/2505.13327",
  year    = "2025",
  month   = may
}

@article{Goodfellow-CACM-2020,
  author  = "Goodfellow, I. and Pouget-Abadie, J. and Mirza, M. and Xu, B. and Warde-Farley, D. and Ozair, S. and Courville, A. and Bengio, Y.",
  title   = "Generative adversarial networks",
  journal = "Communications of the ACM",
  volume  = "63",
  number  = "11",
  pages   = "139--144",
  year    = "2020",
  month   = oct
}

@article{Oquab-TMLR-2024,
  author  = "Oquab, M. and Darcet, T. and Moutakanni, T. and Vo, H. V. and Szafraniec, M. and Khalidov, V. and Fernandez, P. and Haziza, D. and Massa, F. and El-Nouby, A. and Assran, M. and Ballas, N. and Galuba, W. and Howes, R. and Huang, P.-Y. and Li, S.-W. and Misra, I. and Rabbat, M. and Sharma, V. and Synnaeve, G. and Xu, H. and Jegou, H. and Mairal, J. and Labatut, P. and Joulin, A. and Bojanowski, P.",
  title   = "{DINO}v2: {L}earning robust visual features without supervision",
  journal = "Trans. Machine Learning Research",
  year    = "2024",
  month   = jan,
}

@article{Feng-CVPRW-2025,
  author  = "Feng, M. and Ito, K. and Aoki, T. and Ohki, T. and Nishigaki, M.",
  title   = "Leveraging intermediate features of vision transformer for face anti-spoofing",
  journal = "Proc. IEEE/CVF Conf. Comput. Vis. Pattern Recog. Workshops",
  pages   = "3464--3472",
  year    = "2025",
  month   = jun
}

@article{Zhang-CVPR-2019,
  author  = "Zhang, S. and Wang, X. and Liu, A. and Zhao, C. and Wan, J. and Escalera, S. and Shi, H. and Wang, Z. and Li S.Z.",
  title   = "A dataset and benchmark for large-scale multi-modal face anti-spoofing",
  journal = "Proc. IEEE/CVF Conf. Comput. Vis. Pattern Recog.",
  pages   = "919--928",
  year    = "2019",
  month   = jun
}

@article{Zhang-TBIOM-2020,
  author  = "Zhang, S. and Liu, A. and Wan, J. and Liang, Y. and Guo, G. and Escalera, S. and Escalante, H.J. and Li, S.Z.",
  title   = "{CASIA-SURF}: {A} large-scale multi-modal benchmark for face anti-spoofing",
  journal = "IEEE Trans. Biometrics, Behavior, and Identity Science",
  volume  = "2",
  number  = "2",
  pages   = "182--193",
  year    = "2020",
  month   = apr
}

@article{Liu-WACV-2021,
  author  = "Liu, A. and Tan, Z. and Wan, J. and Escalera, S. and Guo, G. and Li, S.Z.",
  title   = "{CASIA-SURF C}e{FA}: {A} benchmark for multi-modal cross-ethnicity face anti-spoofing",
  journal = "Winter Conf. Applications of Comput. Vis.",
  pages   = "1178-1186",
  year    = "2021",
  month   = jan
}

@article{Liu-IFS-2022,
  author  = "Liu, A. and Zhao, C. and Yu, Z. and Wan, J. and Su, A. and Liu, X. and Tan, Z. and Escalera, S. and Xing, J. and Liang, Y. and Guo, G. and Lei, Z. and Li, S.Z. and Zhang, D.",
  title   = "Contrastive context-aware learning for 3{D} high-fidelity mask face presentation attack detection",
  journal = "IEEE Trans. Inf. Forensics Secur.",
  volume = "17",
  pages   = "2497--2507",
  year    = "2022",
  month   = jan
}

@article{Fang-IJCAI-2024,
  author  = "Fang, H. and Liu, A. and Yuan, H. and Zheng, J. and Zeng, D. and Liu, Y. and Deng, J. and Escalera, S. and Liu, X. and Wan, J. and Lei, Z.",
  title   = "Unified physical-digital face attack detection",
  journal = "Proc. Thirty-Third Int'l Joint Conf. Artificial Intelligence",
  pages   = "749--757",
  year    = "2024",
  month   = aug
}

@article{He-CVPRW-2024,
  author  = "He, X. and Liang, D. and Yang, S. and Hao, Z. and Ma, H. and Binjie, M. and Li, X. and Wang, Y. and Yan, P. and Liu, A.",
  title   = "Joint physical-digital facial attack detection via simulating spoofing clues",
  journal = "Proc. IEEE/CVF Conf. Comput. Vis. Pattern Recog. Workshops",
  pages   = "995--1004",
  year    = "2024",
  month   = jun
}
